\documentclass[letterpaper, 10 pt, conference]{ieeeconf}  
\IEEEoverridecommandlockouts                              
\usepackage{xcolor}
\usepackage[pagebackref=false,breaklinks=true,letterpaper=true,colorlinks,bookmarks=false]{hyperref}
\usepackage{amsmath}
\usepackage{algorithm}
\usepackage{algorithmic}
\usepackage{booktabs}
\usepackage{subcaption}
\usepackage{tikz}
\usepackage{textcomp}
\usepackage{hyperref}
\usepackage{lipsum}
\usepackage{amssymb}
\usepackage{graphicx}
\usepackage[font=small]{caption}
\usepackage{cite}

\newcommand\copyrighttext{%
  \footnotesize \textcopyright 2023 IEEE. Personal use of this material is permitted.
  Permission from IEEE must be obtained for all other uses, in any current or future
  media, including reprinting/republishing this material for advertising or promotional
  purposes, creating new collective works, for resale or redistribution to servers or
  lists, or reuse of any copyrighted component of this work in other works.}
\newcommand\copyrightnotice{%
\begin{tikzpicture}[remember picture,overlay]
\node[anchor=south,yshift=10pt] at (current page.south) {\fbox{\parbox{\dimexpr\textwidth-\fboxsep-\fboxrule\relax}{\copyrighttext}}};
\end{tikzpicture}%
}

\title{\LARGE \bf
\textbf{Learning to Solve Tasks with Exploring Prior Behaviours}
}

\author{Ruiqi Zhu$^{1}$, Siyuan Li$^{2}$, Tianhong Dai$^{3}$, Chongjie Zhang$^{4}$, Oya Celiktutan$^{1}$ 
\thanks{$^{1}$ R. Zhu and O. Celiktutan are with the Department of Engineering, King's College London.}
\thanks{$^{2}$ S. Li is with the School of Computer Science and Technology, Harbin Institute of Technology.}
\thanks{$^{3}$ T. Dai is with the Department of Computing Science, University of Aberdeen.}
\thanks{$^{4}$ C. Zhang is with the Institute for Interdisciplinary Information Sciences (IIIS), Tsinghua University.}}% <-this % stops a space

\begin{document}

\maketitle
\copyrightnotice
\thispagestyle{empty}
\pagestyle{empty}

\begin{abstract}
Demonstrations are widely used in Deep Reinforcement Learning (DRL) for facilitating solving tasks with sparse rewards. However, the tasks in real-world scenarios can often have varied initial conditions from the demonstration, which would require additional prior behaviours. For example, consider we are given the demonstration for the task of \emph{picking up an object from an open drawer}, but the drawer is closed in the training. Without acquiring the prior behaviours of opening the drawer, the robot is unlikely to solve the task. To address this, in this paper we propose an Intrinsic Rewards Driven Example-based Control \textbf{(IRDEC)}. Our method can endow agents with the ability to explore and acquire the required prior behaviours and then connect to the task-specific behaviours in the demonstration to solve sparse-reward tasks without requiring additional demonstration of the prior behaviours. The performance of our method outperforms other baselines on three navigation tasks and one robotic manipulation task with sparse rewards. Codes are available at \url{https://github.com/Ricky-Zhu/IRDEC}.
\end{abstract}

\section{Introduction}
Deep reinforcement learning (DRL) has demonstrated impressive performance in sequential decision-making problems, such as video games \cite{mnih2015human, silver2016mastering}, robotics manipulation \cite{singh2019end, zhu2022deep}, and autonomous driving \cite{kiran2021deep}. However, for tasks with sparse rewards, the lack of learning signals can hamper the learning process. To address this, demonstrations are often leveraged to facilitate the learning process. 

In real-world applications, we can often encounter situations under which the initial conditions vary from the demonstration, termed as the \emph{task-specific behaviour demonstration} in this paper, which therefore requires additional prior behaviours to complete the tasks. Prior works require collecting additional demonstration of prior behaviours, which have overlapped states with task-specific behaviour demonstration to overcome the problem \cite{singh2020cog}. However, as the initial conditions can be various, instead of collecting the demonstration of prior behaviours, the robots should be able to adapt to different initial conditions and leverage the available task-specific behaviour demonstration to learn the required prior behaviours. For example, consider we are given the demonstration for the task of \emph{picking up the object from a closed drawer}, but in the training, there are obstacles in the way. Therefore, the robots should be able to explore and learn the essential prior behaviours of \emph{removing the obstacles} first and then conduct the task-specific behaviours of \emph{picking up the object from a closed drawer} to complete the task. 

\begin{figure}[!t]
	\centering
	\vspace*{0.1cm}
	\includegraphics[width=\columnwidth]{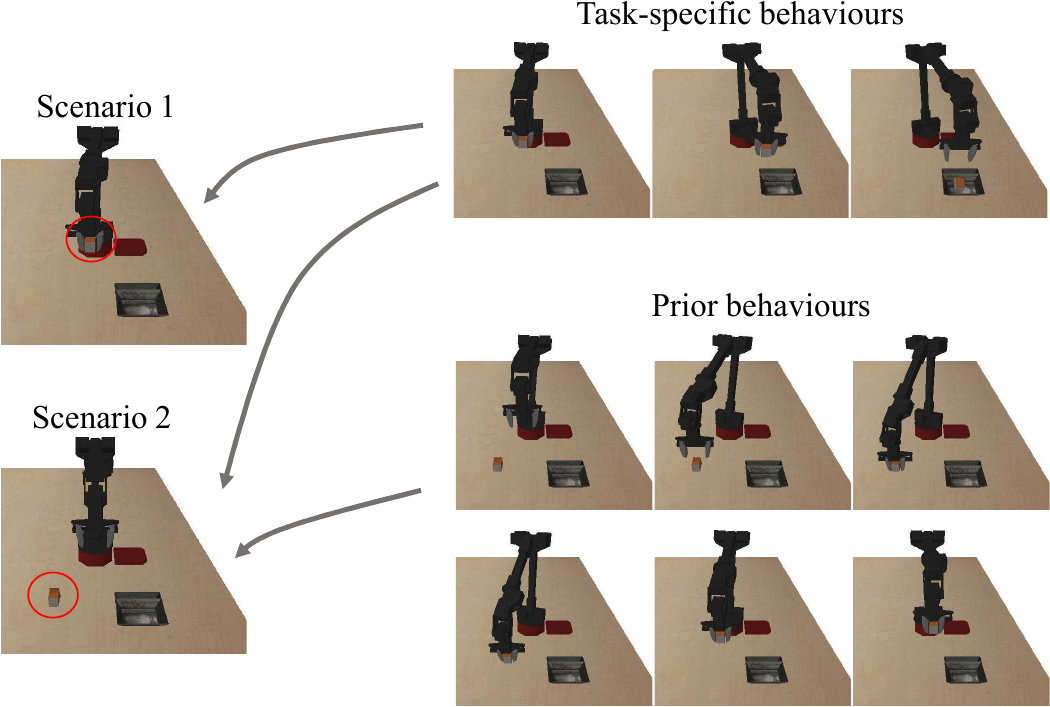}
	\caption{Illustration of the problem definition. Scenario 1: the task of placing the in-hand object into the tray. Scenario 2: the task of placing the object, which is on the table, into the tray. task-specific behaviour demonstration: the demonstration of placing behaviours. Prior behaviours: the behaviours of grasping (top) and lifting (bottom). Scenario 1 can be solved by mimicing the task-specific behaviour demonstration while solving scenario 2 requires the prior behaviours that are absent in the task-specific behaviour demonstration. We attempt to solve scenario 2 given only the task-specific behaviour demonstration without the demonstration of the prior behaviours.}
	\label{fig: motivation}
 \vspace{-0.5cm}
\end{figure}

In this paper, we aim to utilize task-specific behaviour demonstrations to facilitate the learning of tasks with varied initial conditions without requiring additional demonstration of the prior behaviours. The absence of the prior behaviour demonstration indicates that the agent needs to acquire the essential prior behaviours to reach the demonstrated states in the task-specific behaviour demonstration. For example, as illustrated in Fig. \ref{fig: motivation}, the task-specific behaviour is placing the in-hand object into the tray. The agent is required to learn the prior behaviours of grasping and lifting the object from the table, and then mimic the demonstrated actions of the task-specific behaviours to complete the \emph{pick-and-place} task.

When children are shown the task-specific behaviour demonstration and then put in a different initial condition, they would intuitively explore the environment to return to the demonstrated states, which are termed as \emph{familiar states}. Therefore, they can follow the demonstrated actions to complete the tasks. Inspired by that, we propose an intrinsic rewards module to encourage the agents to explore, while the exploration direction is biased towards the familiar states for acquiring the required prior behaviours. Following that, the agents follow the demonstrated actions to complete the tasks, which is achieved with an adaptive behaviour regularizer. The whole framework is trained in an end-to-end manner and can be implemented with off-policy actor-critic RL algorithms, such as soft actor-critic (SAC) \cite{haarnoja2018soft} and deep deterministic policy gradient (DDPG) \cite{lillicrap2015continuous}. Our method was evaluated on challenging long-horizon sparse-reward navigation \cite{li2021active} and robotic manipulation tasks \cite{singh2020cog}. The empirical results show that the proposed method can enable the agent to leverage the task-specific behaviour demonstration to learn the essential prior behaviours to solve tasks with sparse rewards efficiently.

The main contributions of this paper are: (i) We propose an Intrinsic Rewards Driven Example-based Control \textbf{(IRDEC)} learning framework, which enables the agents to acquire prior behaviours and connects to the task-specific behaviours adaptively given only the task-specific behaviour demonstration. (ii) We compare our method with several baselines on challenging navigation and robotic manipulation tasks with sparse rewards and our method achieves the best results. (iii) We carry out an ablation analysis to investigate the importance of each component in our method and the results show that both components, the intrinsic reward module and the example-guided exploration, are necessary for effectively help learn the required prior behaviours.

\section{Related Work}
\textbf{Reinforcement Learning (RL) with Demonstration.} Demonstrations are usually utilized in RL methods to facilitate learning in complex environments or mitigate the problem of sparse rewards. The utilization can often be categorized into: (1) imitation learning in which supervised learning is used to enforce the agent to mimic the demonstrated actions \cite{ross2011reduction} or a reward function is inferred from the demonstration and the policy is trained to optimize it \cite{ho2016generative, li2017infogail, zakka2022xirl}; (2) behaviour extraction in which temporal abstracted behaviours are encoded from a large-scale offline dataset and policy is trained to optimize the extrinsic reward function by acting on the behaviour space to facilitate the exploration \cite{pertsch2021guided, singh2020parrot}; (3) the demonstration is used to initialize the RL actor or regularize the RL actor during the training \cite{rajeswaran2017learning}. However, these works require that the demonstration contains all the required behaviours for the target tasks. But in our proposed method, only the task-specific behaviour demonstration is provided, while the essential prior behaviours are acquired during the training by leveraging the familiar states.

\textbf{Goal-conditioned RL.} Goal-conditioned RL has demonstrated competitive performance on tasks with sparse rewards. By augmenting the state with the goal sampled from the behavioural goal space, the sparse rewards are relabelled with the rewards computed via evaluating the Euclidean distance between the achieved goal and the sampled behaviour goal \cite{andrychowicz2017hindsight, li2021active}. In our proposed method, the task-specific behaviour demonstration contains the goal states of the target tasks, which implicitly implies the goal information. However, goal-conditioned RL requires a mapping function to transform the state space into goal space. In contrast, our method does not require the mapping function and is therefore more applicable.

\textbf{Exploration with Intrinsic Rewards.} In environments with sparse rewards, intrinsic rewards are usually leveraged to encourage the exploration of novel states before any extrinsic rewards are obtained. A large body of works focuses on using forward dynamic model prediction error as intrinsic rewards \cite{pathak2017curiosity, burda2018exploration, nguyen2021sample}. However, intrinsic rewards generated in this way can vanish with the agent being familiar with the environment, and therefore the prediction error tends to converge to zero \cite{badia2020never, raileanu2020ride}. In \cite{ostrovski2017count, seurin2021don}, a count-based exploration bonus is used to incentivize the agent to explore. However, the measure of the counts in large continuous state spaces is non-trivial. Our method incorporates curiosity intrinsic rewards, and impact intrinsic rewards, which do not vanish as the training progresses to encourage the agent to aggressively expand its explored state space so that the familiar states can more easily be encountered.

\begin{algorithm}[tb]
	\caption{Intrinsic Rewards Driven Example-Based Control (IRDEC)}
	\small
	\label{alg:CDECF_algorithm}
	\textbf{Input}:  task-specific behaviour demonstration $\mathcal{D}$
	
	\begin{algorithmic}[1] %[1] enables line numbers
		\STATE \textbf{Initialize:} policy $\pi_{\psi}$, forward dynamics model $f_{\theta_{fw}}$, inverse dynamics model $g_{\theta_{inv}}$, state representation model $\phi_{s}$ , action representation model $\phi_{a}$, classifier $C_{\omega}(s,a)$, intrinsic value head $Q_{\varphi}(s,a)$, online replay buffer $\mathcal{B}\gets \emptyset $, regularization weighting $\lambda_{reg} \gets \lambda_{0}$
		\FOR {$ k = 1, M $}
        \STATE Collect a new trajectory: $\mathcal{B}\gets \mathcal{B}\cup \{\tau\sim\pi_{\psi}\}$

		\STATE Sample transitions $\mathcal{MB}=\{s_{j},a_{j},r_{j}^{e},s'_{j}\}_{j=1}^{K}$ from $\mathcal{B}$
		\STATE Update models within the intrinsic module with samples $\mathcal{MB}$ using Eq. \ref{eq:icm_loss}
		\STATE Compute the intrinsic rewards $r_{j}^{i}$ using Eq. \ref{eq:intrinsic_reward}
		\STATE Replace the rewards $r_{j}^{e}$ in $\mathcal{MB}$ with $r_{j}=r_{j}^{e}+r_{j}^{i}$
		\STATE Optimize the intrinsic value head by minimizing the loss in Eq. \ref{eq:exploration_q_loss}
            \STATE Sample \emph{familiar states} and corresponding demonstrated actions $(s^{*}_{i},a^{*}_{i})$ from  $\mathcal{D}$
		\STATE Optimize the classifier by minimizing the loss in Eq. \ref{eq:classifier_loss}
		\STATE Update the policy using Eq. \ref{eq:policy_update}
		\ENDFOR
	\end{algorithmic}
\end{algorithm}

\section{Background}\label{sec:background}
The learning problem is formulated as a Markov Decision Process (MDP) characterized by a tuple $\{\mathcal{S, A, T, R,\rho,\gamma}\}$ of state space, action space, transition probability mapping from current state $s$ and action $a$ to the next state $s'$, reward function, initial state distribution, and discount factor. In each episode, the initial state $s_{0}\in\mathcal{S}$ is sampled from the initial state distribution $s_{0}\sim \mathcal{\rho}(s_{0})$, the agent chooses its action $a_{t}\in \mathcal{A}$ according to the policy $a_{t}\sim \pi(\cdot|s_{t})$, and then the environment will generate the next state $s_{t+1}\sim \mathcal{T}(\cdot|s_{t},a_{t})$ and the reward $r=\mathcal{R}(s_{t},a_{t},s_{t+1})$. The objective of the agent is to maximize the sum of discounted rewards, $\mathbb{E}_{\tau\sim\pi}\left[\sum_{t=0}^{t=\infty}\gamma^{t}r_{t}\right]$, where trajectory $\tau$ is sampled from the policy $\pi$.

\textbf{Example-based Control.} 
Unlike the goal-conditioned RL, which requires the mapping from the state space to the goal space, example-based control provides another way of reaching the goal over the future state distribution, defined as Eq. \ref{eq:future state dist} where $s_{t+}$ is the future state. Given examples of goal states, the actor policy is optimized to maximize the probability of reaching these states \cite{eysenbach2021replacing,eysenbach2020c}. 

\begin{align} \label{eq:future state dist}
    p^{\pi}(s_{t +}|s_{t},a_{t}) \triangleq (1-\gamma) \sum_{\Delta=0}^{\infty}\gamma^{\Delta}p^{\pi}(s_{t+\Delta}=s_{t +}|s_{t},a_{t})
\end{align}
The probability as shown in Eq. \ref{eq:exm_task_solve_prob} is defined over the future state distribution.

\begin{align} \label{eq:exm_task_solve_prob}
    p^{\pi}(e_{t +}|s_{t},a_{t}) = \mathbb{E}_{p^{\pi}(s_{t +|s_{t},a_{t}})}[p(e_{t +}|s_{t +})]
\end{align}
where $e_{t +}$ is referred to as the event of reaching the example states conditioned on the future state from time step $t$. In our method, we leverage the classifier in \cite{eysenbach2021replacing} to estimate the probability of reaching the demonstrated states in the task-specific behaviour demonstration in the future. However, the exploration introduced by maximizing the classifier values could often encounter danger areas in tasks, which can result in failure to solve the tasks. Please refer to Section~\ref{section:intrinsic_reward} for further details.  

\textbf{Off-policy Actor-critic RL.} Our method can be implemented with off-policy actor-critic algorithms. In these algorithms, the critic learns an off-policy estimate of the value function for the current actor policy with the samples collected from behavioural actor policies \cite{degris2012off}. The value function in our method estimates the weighted sum of the future return of the extrinsic and intrinsic rewards and the classifier values under the target actor policy. The actor policy in turn is optimized to maximize the estimates using the off-policy samples which are collected online.
% To achieve that, a classifier is used to predict the probability, which is trained with the success examples and online collected transitions.

%%%%%%%%%%%%%%%%
\section{Intrinsic Rewards Driven Example-Based Control Framework}
\label{section:intrinsic_reward}
\begin{figure}[!t]
	\centering
	\vspace*{0.1cm}
	\includegraphics[width=\columnwidth]{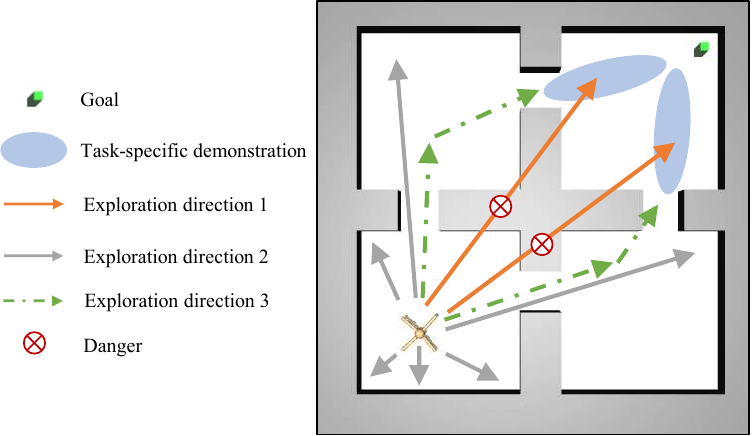}
	\caption{Illustration of the proposed method. Exploration direction 1: the exploration direction introduced by the classifier. Exploration direction 2: the exploration direction introduced by the intrinsic reward module. Exploration direction 3: the exploration direction introduced by our method.}
	\label{fig: algorithm box}
  \vspace{-0.5cm}

\end{figure}

The proposed method consists of two main components, namely, a curiosity-impact driven intrinsic reward module that encourages the agent to expand explored areas, and example-guided exploration provided by a classifier that predicts the probability of reaching familiar states in the future. 
To reach the familiar states and connect the task-specific behaviours, proper exploration over the state space is essential.
The exploration direction introduced by directly maximizing the classifier values without the intrinsic reward module can potentially encounter danger areas. The exploration introduced by the intrinsic rewards module without the guidance provided by the classifier can cover less promising areas related to the target tasks, which can be inefficient and could hinder the agent from solving the tasks. 
By expanding the explored areas towards the familiar states with leveraging the guidance from the task-specific behaviour demonstration, our proposed method can learn the essential prior behaviours and connect them to the behaviours in the task-specific behaviour demonstration for solving the target tasks
as illustrated in Fig. \ref{fig: algorithm box}.

\subsection{Curiosity-Impact Driven Intrinsic Reward Module}
Our intrinsic reward module combines curiosity intrinsic rewards that encourage the agent to visit novel states \cite{pathak2017curiosity} and impact intrinsic rewards that incentivize the agents to explore local states with large differences from current states, which can help to reach critical stages during the exploration \cite{zhang2020bebold}. We train a forward and an inverse dynamics models to learn the state representation $\phi_{s}(s)$ and the action representation $\phi_{a}(a)$. The forward dynamics model parameterized by $\theta_{fw}$ is used to predict the representation of the next state $\phi_{s}(s_{t+1})$ given the current state representation $\phi_{s}(s_{t})$ and action representation $\phi_{a}(a_{t})$. The loss for the forward dynamics model is given in Eq. \ref{eq:forward_loss}:

\begin{align} \label{eq:forward_loss}
    L_{fw}(s_{t},a_{t},s_{t+1}) = \frac{1}{2}\lVert  f_{\theta_{fw}}(\phi_{s}(s_{t}),\phi_{a}(a_{t})) - \phi_{s}(s_{t+1})  \rVert_{2}^{2}
\end{align}

The inverse dynamics model parameterized by $\theta_{inv}$ is used to predict the action representation $\phi_{a}(a_{t})$ given the consecutive state representations $\phi_{s}(s_{t})$ and $\phi_{s}(s_{t+1})$. The loss for the inverse dynamics model is shown in Eq. \ref{eq:inverse_loss}. With the inverse dynamics model, the adverse impact on state representation learning, caused by the inherent noise of the environment where transition might not be affected by the agent's actions, i.e. noisy TV \cite{ostrovski2017count, burda2018exploration},  can be mitigated by retrieving the action leading to the transition.

\begin{align} \label{eq:inverse_loss}
    L_{inv}(s_{t},a_{t},s_{t+1}) = -\log (g_{\theta_{inv}}(\phi_{a}(a_{t})|\phi_{s}(s_{t}),\phi_{s}(s_{t+1}))
\end{align}

Thus, the total loss for the intrinsic module can be written as follows:
\begin{align} \label{eq:icm_loss}
    L_{icm} =  L_{fw} + L_{inv}
\end{align}

 The curiosity intrinsic reward is defined in Eq. \ref{eq:curiosity_reward}. A larger curiosity intrinsic reward, namely a larger prediction error, indicates the novel states being visited as the forward dynamics model is unfamiliar with the transition.
\begin{align} \label{eq:curiosity_reward}
    r_{t}^{curiosity} = \lVert  f_{\theta_{fw}}(\phi_{s}(s_{t}),\phi_{a}(a_{t})) - \phi_{s}(s_{t+1})  \rVert_{2}
\end{align}

However, as the training progresses, the curiosity intrinsic rewards could vanish as the agent is being familiar with the environment. Thus, the impact intrinsic rewards defined as the squared Euclidean distance between consecutive state representations, as shown in Eq. \ref{eq:intrinsic_impact__reward}, are utilized to encourage the agent to aggressively change its states to accelerate the exploration.
\begin{align} \label{eq:intrinsic_impact__reward}
    r_{t}^{impact} =  \frac{\lVert  \phi_{s}(s_{t+1})-\phi_{s}(s_{t})  \rVert_{2}}{d_{m}^{2}}, 
\end{align}
where $d_{m}$ is the running average of the numerator for scaling the bonus. The impact intrinsic rewards will not vanish as the training progresses.
Overall, the intrinsic reward can be written as:

\begin{align} \label{eq:intrinsic_reward}
    r_{t}^{i} = \eta r_{t}^{curiosity}+(1-\eta)r_{t}^{impact},
\end{align}
where $\eta>0$ is a scalar weighing the two types of intrinsic rewards. Thus, the overall reward $r_{t}$ at step $t$ is defined as the addition of intrinsic reward $r_{t}^{i}$ and the sparse extrinsic reward $r_{t}^{e}$.
% \begin{align}
%     r_{t} = r_{t}^{i}+r_{t}^{e}
% \end{align}
A value head parameterized by $\varphi$ is trained to approximate the action value based on the overall reward. We optimize the value head by minimizing the loss:

\begin{align}\label{eq:exploration_q_loss}
    L_{Q} = (r_{t}+\gamma \mathbb{E}_{a_{t+1}\sim\pi}Q_{\varphi^{'}}^{\pi}(s_{t+1},a_{t+1})-Q_{\varphi}^{\pi}(s_{t},a_{t}))^{2},
\end{align}
where $Q_{\varphi^{'}}^{\pi}$ is the target network of the value head $Q_{\varphi}^{\pi}$ for stabilizing the training \cite{mnih2015human}.

\subsection{Example-guided Exploration}
To learn the essential prior behaviours and connect them to the task-specific behaviours, visiting the overlapped state distribution of these behaviours are necessary \cite{singh2020cog}. Here we encourage the agent to visit the familiar states in the task-specific behaviour demonstration to construct such overlapped state distribution. To enable the agent to reach the familiar states, we utilize the classifier in \cite{eysenbach2021replacing} to discriminate between the state-action pairs which lead to reaching the familiar states in the future or not. The positive state-action pair is defined as the pair of familiar states sampled from task-specific behaviour demonstration $\mathcal{D}$ and the action given by the current policy conditioned on the sampled familiar states. The negative state-action pairs are those sampled from the online buffer $\mathcal{B}$. Thus, the relation between the optimal classifier and $p^{\pi}(e_{t +}|s_{t},a_{t})$ can be written as:
\begin{align}
    C_{\omega}(s_{t},a_{t})=\frac{p^{\pi}(s_{t},a_{t}|e_{t+}=1)p(e_{t+}=1)}{p^{\pi}(s_{t},a_{t}|e_{t+}=1)p(e_{t+}=1)+p(s,a)}.
\end{align}
Thus, the objective to optimize can be derived as:
\begin{align}
    p^{\pi}(e_{t+}=1|s_{t},a_{t}) = \frac{C_{\omega}(s_{t},a_{t})}{1-C_{\omega}(s_{t},a_{t})}.
\end{align}
The loss for the classifier is shown as Eq. \ref{eq:classifier_loss}. In the equation, $\mathcal{CE}$ denotes the cross entropy loss.  

\begin{equation}
    \label{eq:classifier_loss}
    \begin{aligned}
    \mathcal{L}(\theta)=&(1-\gamma)\mathbb{E}_{s\sim \mathcal{D},a\sim \pi(\cdot|s)}\mathcal{CE}(C_{\omega}(s,a);y_{pos})\\
    &+(1+\gamma w)\mathbb{E}_{(s,a,s')\sim B}\mathcal{CE}(C_{\omega}(s,a);y_{neg})
\end{aligned}
\end{equation}

The positive target $y_{pos}$ is $1$, while the negative target $y_{neg}$ is computed as follows:

\begin{align}
    y_{neg}=\frac{\gamma w(s')}{1+\gamma w(s')},
\end{align}
where the $w(s)$ is computed as shown in Eq. \ref{eq:w}. The detailed derivation can be referred to \cite{eysenbach2021replacing}.

\begin{align}\label{eq:w}
    w(s) = \frac{C_{\omega}^{\pi}(s,\pi(\cdot|s))}{1-C_{\omega}^{\pi}(s,\pi(\cdot|s))} 
\end{align}

\subsection{Adaptive Behaviour Regularization}
During the training, the agent should maintain the capacity of executing task-specific behaviours after encountering the familiar states, i.e., after grasping the object the robot needs to know how to place the object into the tray by mimicing the demonstrated actions in the task-specific behaviour demonstration. Here, we utilized Eq. \ref{eq:reg} as the regularization loss to the update of the actor.
\begin{align}\label{eq:reg}
    L_{reg}(s^{*},a^{*}) = -\log(\pi_{\psi}(a^{*}|s^{*})),
\end{align}
where $(s^{*}, a^{*})$ are the state-action pairs sampled from the task-specific behaviour demonstration.
\begin{figure*}
	\centering
	\includegraphics[width=\textwidth]{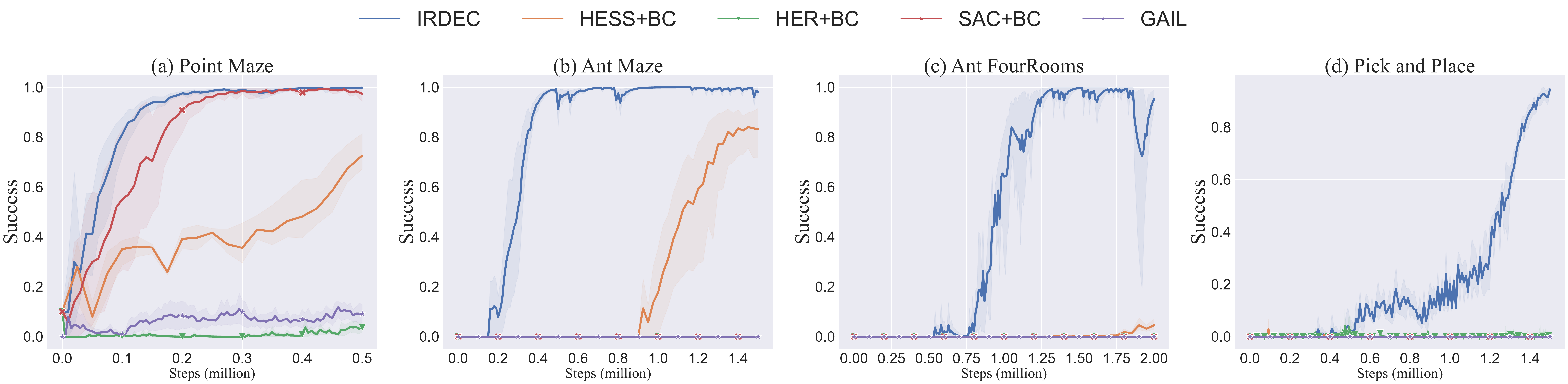}
	\small
	\caption{Learning curves of the proposed method and baselines on all tasks. All the curves are smoothed equally for visual clarity.}
	\label{fig:performance graph}
  \vspace{-0.5cm}
\end{figure*}
However, the weighting of the regularization in the actor update should vary in different stages of the training. In the early stage of the training, the agent focuses on exploring and reaching the familiar states; therefore, the weighting should be relatively small to avoid the adverse impact on the exploration of familiar states. When the agent acquires the prior behaviours to reach the familiar states, the agent should focus more on exploiting the task-specific behaviour demonstration and mimicing the demonstrated actions, where the weighting should be relatively large. The similarity between the sampled online collected states and familiar states increases as the agent is more capable of reaching the familiar states. Thus, we leverage a kernel density estimator fitted with the familiar states in the task-specific behaviour demonstration to estimate the similarity and therefore adjust the regularization of loss weighting. However, as the range of estimation scores is unknown, the effective way to adjust the weighting is to compare the scores of consecutive sampled batches. The weighting is computed as:

\begin{equation}
    \lambda_{reg} = \text{clip}(\lambda_{i-1} + \frac{m(b_{i})-m(b_{i-1})}{\max(m(b))}\times r, \lambda_{min}, \lambda_{max})
\end{equation}
where the $\lambda_{i-1}$ is the weighting value at update step $i-1$, and $m(b_{i})$ is the density estimation score of the sampled batch of states at update step $i$. r is the pre-defined rate. $\max(m(b))$ and $\lambda_{0}$ are the recorded maximum estimation score and the initial value of $\lambda_{reg}$. Meanwhile, we clip the $\lambda_{reg}$ which is out of the pre-defined range $[\lambda_{min}, \lambda_{max}]$.

Overall, the parameters of the actor policy network can be updated as follows: 
\begin{equation}
    \begin{aligned}\label{eq:policy_update}
    \psi \gets \psi &+ \lambda\nabla_{\psi}\mathbb{E}_{a_{t}\sim \pi_{\psi}, s_{t}\sim \mathcal{B}}[(C(s_{t},a_{t})+Q(s_{t},a_{t}))] \\
    &+\lambda_{reg}\nabla_{\psi}\mathbb{E}_{(s^{*},a^{*})\sim\mathcal{D}}L_{reg}(s^{*},a^{*}).
\end{aligned}
\end{equation}

% In each update, the state-action pairs are sampled from task-specific behaviour demonstration $\mathcal{D}$ and online buffer $\mathcal{B}$ respectively to update the classifier $C_{\omega}$ and the value head $Q_{\varphi}$. Then the actor $\pi_{\psi}$ is updated by maximizing the classifier and the value head based on an off-policy actor-critic method. The intrinsic module in our method encourages the agent to explore while maximizing the classifier values to steer the exploration direction towards reaching the familiar states. Following that, adaptive regularization is utilized to enforce the actor policy mimicing the demonstrated actions to solve the tasks.

\begin{figure}[!t]
	\centering
	\vspace*{0.1cm}
	\includegraphics[width=\columnwidth]{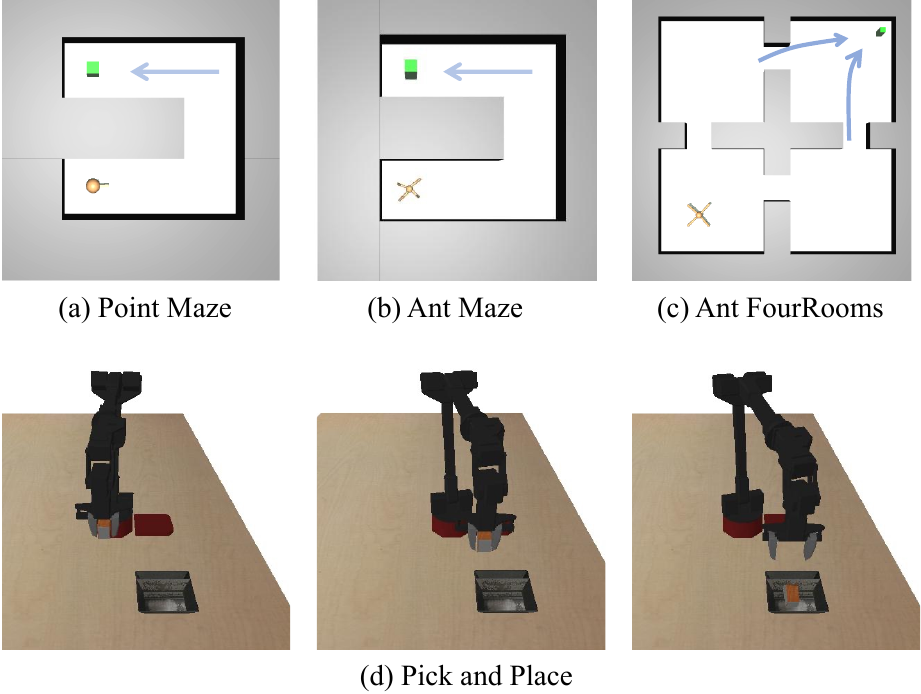}
	\caption{Illustration of settings for all the tasks. (Top) Three navigation tasks: (a) Point Maze, (b) Ant Maze, and (c) Ant FourRooms. The arrows represent the trajectories of the task-specific behaviour demonstration. (Bottom) Pick and Place task. The sequence of the three frames represents the task-specific behaviour demonstration (placing behaviours).}
	\label{fig:four_environments}
  \vspace{-0.5cm}
\end{figure}

\section{experiment} \label{sec:experiment}
We aim to answer the following questions through our experiments: \textbf{(1)} Can our proposed method effectively and efficiently leverage the task-specific behaviour demonstration to learn the essential prior behaviours and further develop a complete policy for solving the target tasks with sparse rewards? \textbf{(2)} How does our method compare to alternative methods? \textbf{(3)} What is the importance of each component of our method?

\subsection{Experiment setup}
We evaluate the proposed framework on three long-horizon navigation tasks and one robotic manipulation task as illustrated in Fig. \ref{fig:four_environments}. In our problem settings, we assume the access to the task-specific behaviour demonstration $\mathcal{D}$ in the form of state-action trajectories $\tau_{i}=\{(s^{*}_{0},a^{*}_{0}, ..., s^{*}_{T_{i}}, a^{*}_{T_{i}})\}$. In our experiments, the task-specific behaviour demonstration for each task contains $100$ trajectories collected with a sub-optimal policy. The trajectory lengths are around $150$ and $25$ for the navigation tasks and the robotic manipulation task respectively.

\noindent \textbf{Navigation Tasks.} We evaluated our method on three simulated long-horizon navigation tasks with sparse rewards based on Mujoco \cite{todorov2012mujoco}. In these tasks, the agent should learn to control the robot (point or ant) to navigate through the maze to reach the goal as illustrated in Fig. \ref{fig:four_environments} (a-c). For the point robot, the observation space $\mathcal{O} \in \mathbb{R}^{6}$ consists of the positions and velocities of the mass centre, and action space $\mathcal{A}\in \mathbb{R}^{2}$. For the ant robot, the observation space $\mathcal{O} \in \mathbb{R}^{29}$ consists of the positions and velocities of its torsos, and action space $\mathcal{A}\in \mathbb{R}^{8}$. The extrinsic reward is given as $+1$ when the robot reaches the goal and $0$ otherwise. The task-specific behaviour demonstration for each task is illustrated as arrows in Fig. \ref{fig:four_environments} (a-c). The agent is supposed to learn the essential prior behaviours of navigating the robot to the familiar states and then conduct the behaviours of the task-specific behaviour demonstration to solve the tasks.
\begin{figure*}
	\centering
	\includegraphics[width=\textwidth]{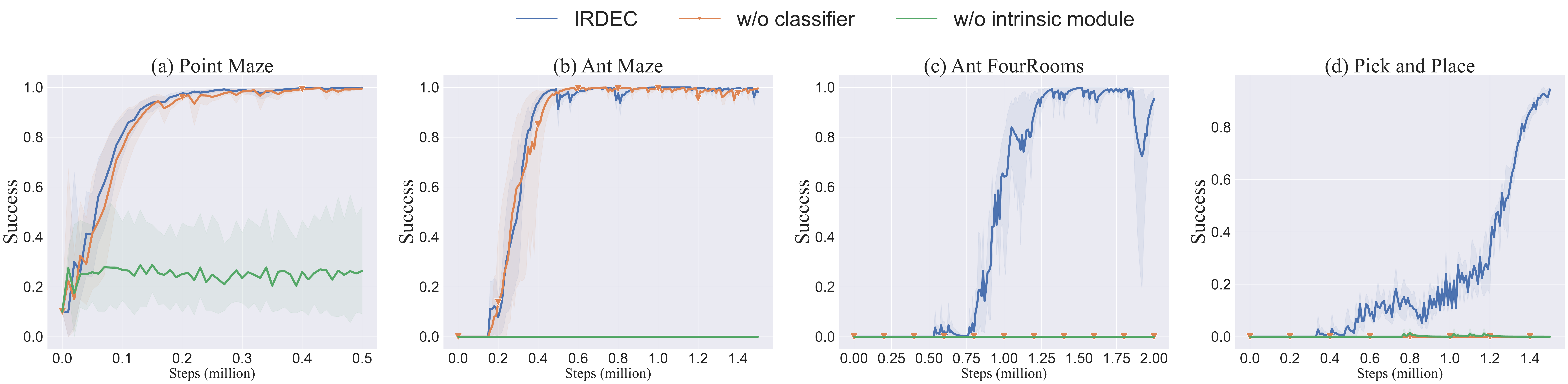}
	\small
	\caption{Ablation studies of our proposed method.}
	\label{fig:ablation graph}
  \vspace{-0.5cm}
\end{figure*}
\begin{figure}[!t]
	\centering
	\vspace*{0.1cm}
	\includegraphics[width=\columnwidth]{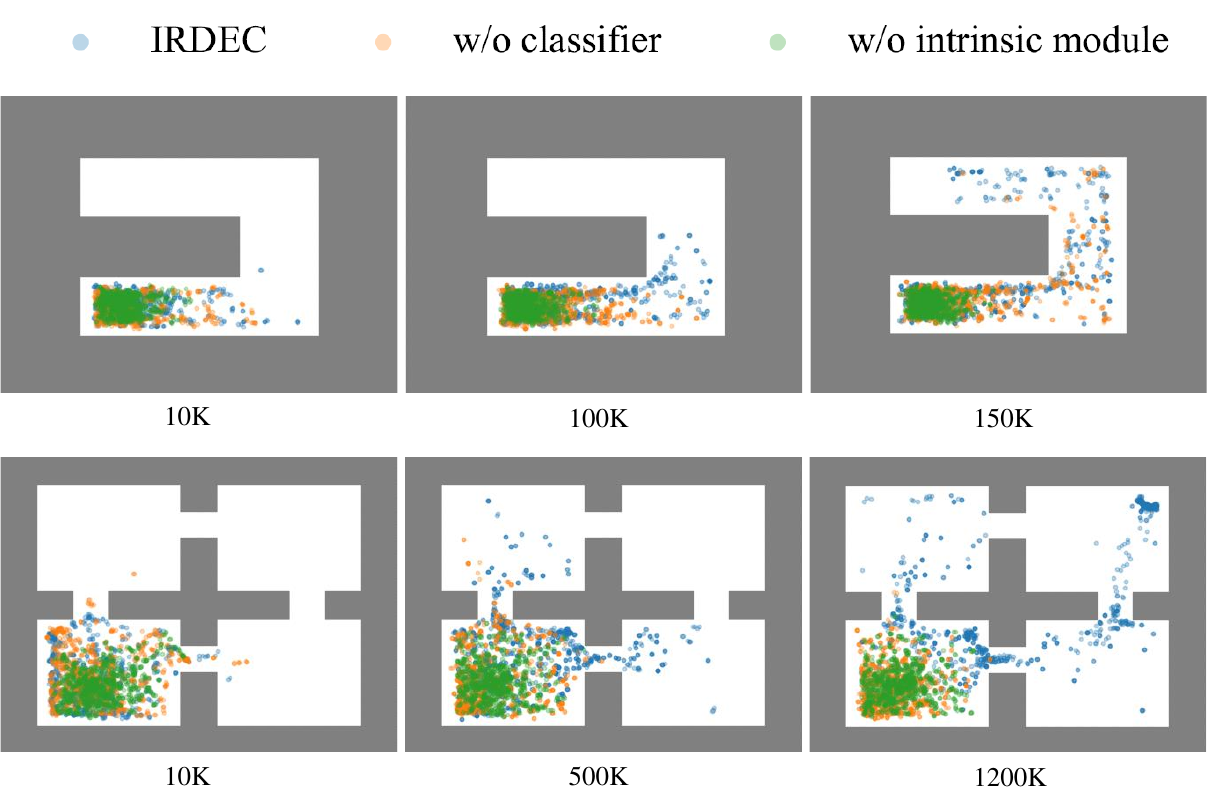}
	\caption{Visulization of explored areas. 1K points are sampled uniformly from the online buffer to represent the explored areas. Top: the explored areas after training steps of 10K, 100K, and 150K for the Ant Maze task. Bottom: the explored areas after training steps of 10K, 500K, and 1200K for the Ant FourRooms task.}
	\label{fig:explored_area}
  \vspace{-0.5cm}
\end{figure}

\noindent \textbf{Pick and Place Task.} 
To evaluate our method on robotic manipulation tasks, we utilized the Pick and Place environment in \cite{singh2020cog}. The simulated environment consists of a 6-DoF Widow X robot in front of a tray. The robot is supposed to learn how to control its arm to grasp the object from the table and place it in the tray. The observation space $\mathcal{O}\in \mathbb{R}^{17}$, includes the state of the end-effector and the gripper. The action space $\mathcal{A} \in \mathbb{R}^{8}$, includes the Cartesian coordinate changes, orientation changes of the end-effectors, and the gripper open degree. In the task, the agent needs to learn how to \emph{grasp}, \emph{lift} the object, and \emph{place} the object in the tray. The extrinsic reward is $+1$, when the objective is placed in the tray and otherwise $0$. For this task, the task-specific behaviour demonstration consists of the trajectories of controlling the robot to place the in-hand object into the tray as illustrated in Fig. \ref{fig:four_environments} (d). The robot needs to learn the essential prior behaviours of grasping and lifting the object from the table and connect them to the task-specific behaviour to solve the task.

\noindent \textbf{Baselines.} We compare our method with: \textbf{(1) HER+BC} \cite{andrychowicz2017hindsight}, a goal-conditioned method that utilizes a goal relabelling method to augment sufficient positive samples in goal-conditional tasks, while applying behaviour cloning loss to regularize the actor update with the task-specific behaviour demonstration. \textbf{(2) HESS+BC} \cite{li2021active}, a goal-conditioned hierarchical method which utilized the high-level policy to assign sub-goal for the low-level policy to facilitate exploration, which has demonstrated competitive performance in long-horizon tasks with sparse rewards. Meanwhile, behaviour cloning loss is applied to regularize the actor update with the task-specific behaviour demonstration. \textbf{(3) SAC+BC} \cite{haarnoja2018soft}, which trains the agent with SAC while applying behaviour cloning loss to regularize the actor update with the task-specific behaviour demonstration. \textbf{(4) GAIL} \cite{ho2016generative}, an imitation learning method which attempts to match the state distribution between the training data and the demonstration. The hyperparameters used in these baseline methods were adopted from the original papers.

\subsection{Experimental results}
During the training, we evaluated the performance every 10K training steps with 10 episodes and recorded the average test returns. For each task, we trained our method and baselines with 5 different seeds and reported the results in Fig.~\ref{fig:performance graph}. As shown in the figure, the proposed method outperforms all the baselines in all tasks. The outperformance is more significant in the Ant FourRooms task and the Pick and Place task as the exploration problems are harder. Without an effective incentive for expanding explored area, the sampled behavioural goals of HER+BC are insufficient with respect to diversity, which accounts for its poor performance in these long-horizon tasks \cite{pitis2020maximum}. By leveraging the hierarchical structure, HESS+BC assigns informative behavioural subgoals to encourage the low-level policy to visit promising areas for solving the tasks. However, it presents less efficiency compared to our proposed method in all the long-horizon navigation tasks and it fails in the Pick and Place task as the subgoal representation is not well learned in the robotic manipulation task. SAC+BC performs poorly in all tasks except for the Point Maze task as the action space and observation space are relatively low-dimensional so the entropy term in the actor update is sufficient for addressing the exploration problem. GAIL failed in all tasks because the policy cannot generate the training data that matches the demonstration.

\subsection{Ablation analysis}
To understand the role and importance of each component in our method, we conducted an ablation analysis on the curiosity-impact driven intrinsic reward module and the example-guided exploration separately. We compare IRDEC, IRDEC without the intrinsic rewards module, and IRDEC without the classifier. As shown in Fig. \ref{fig:ablation graph}, our method without the intrinsic rewards module failed in all the tasks while our method without the classifier can solve the Point Maze and the Ant Maze tasks but failed in the hard Ant FourRooms and Pick and Place tasks. To understand the underlying causes, we visualize the explored areas for the tasks of the Ant Maze and the Ant FourRooms as shown in Fig. \ref{fig:explored_area}. We sampled 1K points from the online buffer to represent the explored areas after specific training steps. IRDEC without the intrinsic reward module failed both tasks, as the exploration attempts to pass through the wall directly towards the familiar states. And in robotic manipulation tasks, Pick and Place, the "wall" could be the unachievable robot configurations due to the singularity.
For the Ant Maze task, IRDEC and IRDEC without the classifier can expand their explored areas towards the goal while IRDEC is slightly more efficient as the exploration direction is biased towards familiar states. For the Ant FourRooms task, only IRDEC succeed in reaching the goal. IRDEC without the classifier lacks guidance towards the goal and therefore the exploration directions are random. When the state spaces are large, the random exploration direactions could lead the failure in passing the critical points, such as the narrow doors connecting each room in Ant FourRooms. Moreover, As the training progresses, the curiosity intrinsic rewards part in the intrinsic reward module vanishes, which makes the exploration harder. The ablation analysis demonstrates that IRDEC can effectively leverage the familiar states in the task-specific behaviour demonstration to bias the exploration direction introduced by the intrinsic reward module for learning the required prior behaviours and connecting them to the task-specific behaviour to solve the tasks.

\section{Conclusion}
In this paper, We present IRDEC, a method that incorporates an intrinsic reward module to proactively expand explored areas while biasing the exploration towards familiar states in the task-specific behaviour demonstration, for endowing the agent with the capabilities to adapt to initial conditions that are unseen from the demonstration. Our proposed method shows its capability of automatically learning the essential prior behaviours and connecting them to the behaviours in the task-specific behaviour demonstration for solving long-horizon tasks with sparse rewards. With our method, agents can adapt to tasks with varied initial conditions from the task-specific behaviour demonstration without requiring additional demonstration of the required prior behaviours. Additionally, the empirical results show the efficiency and viability of IRDEC compared to all the baselines. An exciting direction for future work would be applying the method to high-dimensional pixel-based control. %Another promising direction would be to further facilitate the exploration towards the task-specific behaviour demonstration to make it more practical to apply the method to real-world robotic tasks.

%\newpage
\bibliographystyle{IEEEtran}
\bibliography{main}

\end{document}